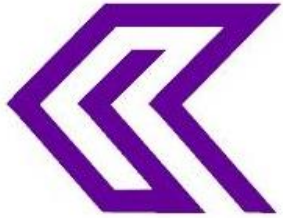

# CLUSTERING APPROACH TOWARDS IMAGE SEGMENTATION: AN ANALYTICAL STUDY

**Dibya Jyoti Bora[1], Dr. Anil Kumar Gupta[2]**

[1] Department of Computer Science and Applications, Barkatullah University, Bhopal, India,
research4dibya@gmail.com, mobile no. 9074966860

[2] HOD, Department of Computer Science and Applications, Barkatullah University, Bhopal, India,
akgupta_bu@yahoo.co.in

DEPARTMENT OF COMPUTER SCIENCE & APPLICATIONS, BARKATULLAH UNIVERSITY, BHOPAL-462026

**Abstract:**

Image processing is an important research area in computer vision. Image segmentation plays the vital rule in image processing research. There exist so many methods for image segmentation. Clustering is an unsupervised study. Clustering can also be used for image segmentation. In this paper, an in-depth study is done on different clustering techniques that can be used for image segmentation with their pros and cons. An experiment for color image segmentation based on clustering with K-Means algorithm is performed to observe the accuracy of clustering technique for the segmentation purpose.

## 1. Introduction:

For image processing research, image segmentation always acts as major research topic. Image segmentation is needed to detect objects or divide the image into regions which can be considered homogeneous according to a given criterion, such as color, motion, texture, etc. As an essential component of an image analysis and or pattern recognition system, image segmentation always draws researchers' attention for developing new and efficient technique for the segmentation purpose. There always arises the need of a novel technique for image segmentation purpose as because this is one of the most critical tasks in image processing. Clustering is an unsupervised study. This unsupervised study is done to separate data objects into different groups, known as clusters so that within one cluster, objects with same properties belong and they differ from objects in another clusters. In this paper, an analytical study is done on different clustering techniques that can be applied for image segmentation purpose. An experiment is done at the end to see the quality of segmentation task through clustering for which we have chosen K-Means algorithm as the basic clustering algorithm.

## 2. Image Segmentation:

An image is not just a random collection of pixels; it is a meaningful arrangement of regions and objects. Extracting information from an image is referred to as image analysis which is one of the preliminary steps in pattern recognition systems. Each region of the image is made up of set of pixels. Image segmentation can be defined as the classification of all the picture elements or pixels in an image into different clusters that exhibit similar features [1]. This is a process of subdividing an image into its constituent's parts or objects in the image i.e. set of pixels such that pixels in a region are similar considering some homogeneity criteria such as color, intensity or texture so as to locate and identify boundaries in an image [2]. An image is segmented to further analyze each of these objects present in the image to extract some high level information. The result of image segmentation is a set of segments that collectively cover the entire image, or a set of contours extracted from the image [3]. Adjacent regions are significantly different with respect to the same characteristic(s). Segmentation has been used in a wide range of applications. Different applications require different types of images. The most commonly used images are light intensity (LI), range (depth) image (RI), computerized tomography (CT), magnetic resonance images (MRI). Image segmentation is highly dependent on the image type; hence there is no single generalized technique that is suitable for all images.

Normally, images are divided into two types on the basis of their color, i.e. gray scale and color images. Therefore image segmentation for color images is totally different from gray scale images, e.g., content based image retrieval [4][5].Also, segmentation can be either local or global [6]. Local segmentation is small windows on a whole image and deals with segmenting sub image. Global segmentation deals with segmenting whole image. Global segmentation mostly deals with relatively large no of pixel. But local segmentation deal with lower no of pixel as compare to global segmentation.

Image segmentation methods can be broadly classified into seven groups[7][8]: (1) Histogram thresholding, (2) Clustering (Soft and Hard), (3) Region growing, region splitting and merging, (4) Edge-based, (5) Physical model-based, (6) Fuzzy approaches, and (7) Neural network and GA (Genetic algorithm) based approaches.

In this paper, we will discuss different techniques for image segmentation based on clustering.

## 3. Different clustering algorithms for image segmentation:

Clustering is a very important area of data mining which has applications in almost every field of science and engineering. This is an unsupervised study where we try to partition a given set of data items into different subparts known as clusters. But on doing this we have to maintain two criteria: 1. High cohesive property, & 2.Low coupling property.

Clustering can be used for image segmentation tasks. In clustering, we can approach either hard wise (known as "hard clustering" or "exclusive clustering") or soft wise("soft clustering" or "overlapping clustering")[9]. In case of hard clustering, a data item can belong to only one cluster; while in case of soft clustering a data item may belong to more than one cluster.

## 4. Hard Clustering Algorithms for image segmentation:

Following are some frequently used hard clustering algorithms that can be used for image segmentation:

*4.1 Naïve K-Means:*

**K-means algorithm** is also referred to as Lloyd's algorithm. This is an unsupervised clustering algorithm that classifies the input data points into multiple classes based on their inherent distance from each other. The algorithm assumes that the data features form a vector space and tries to find natural clustering in them. The points are clustered around centroids which are obtained by minimizing the objective function. In this algorithm, we try to minimize the empirical mean of squared distance between data points and cluster centroids.

The main steps involved in this algorithm are [10]:
1. Select an initial partition with k clusters
2. Generate a new partition by assigning each pattern to its closest cluster center.
3. Compute new cluster centers.
4. Continue to do steps 2 and 3 until memberships finalize.

K-Means algorithm [11][12][13][14] aims at minimizing an *objective function*, in this case a squared error function. The objective function is:

$$J = \sum_{j=1}^{k} \sum_{i=1}^{n} \left\| x_i^{(j)} - c_j \right\|^2$$

Where $\left\| x_i^{(j)} - c_j \right\|^2$ is a chosen distance measure between a data point $x_i^{(j)}$ and the cluster centre $c_j$, is an indicator of the distance of the *n* data points from their respective cluster centers. But choosing an accurate distance measure is very important as it has direct effects on the clustering tasks [15].

Though K-Means algorithm is simple, but here a critical issue is that initialization of centers should be done very carefully. Also choosing a proper value of k(number of clusters) is utmost necessary for better performance of the algorithm.

### *4.2 Accelerated K-Means:*

This is an optimized version of K-Means algorithm [16]. This is based on the fact that most distance calculations in standard K-Means are redundant. Suppose a point is far away from a center, then it is not necessary to calculate the exact distance between the point and the center in order to know that the point should not be assigned to this center. This algorithm applies the triangle inequality in two different ways, and by keeping track of lower and upper bounds for distances between points and centers in order to avoid unnecessary distance calculations. Experiments has also been done which show that the new algorithm is effective for datasets with up to 1000 dimensions, and becomes more and more effective as the number *k* of clusters increases[11][13].

### *4.3 K-Means++:*

K-Means++ is a solution to the initialization problem of K-Means algorithm. This method can be used to initialize the number of cluster *k* and then this k is given as an input to the *k*-means algorithm. Since choosing the right value for *k* in prior is difficult, this algorithm provides a method to find the value for *k* before proceeding to cluster the data[17]. The exact algorithm is as follows:
1. Choose one center uniformly at random from among the data points.
2. For each data point *x*, compute D(*x*), the distance between *x* and the nearest center that has already been chosen.
3. Choose one new data point at random as a new center, using a weighted probability distribution where a point *x* is chosen with probability proportional to D(*x*)$^2$.
4. Repeat Steps 2 and 3 until *k* centers have been chosen.
5. Now that the initial centers have been chosen, proceed using standard *k*-means clustering.

### 4.4 K-Medoids:

This algorithm clusters data points based on closest center like the *k*-means algorithm. Unlike *k*-means algorithm, which replaces each center by the mean of points in the cluster, *k*-medoids replaces each center by the Medoids of points in the cluster. Medoid is the centrally located data point of the cluster. Therefore in *k*-medoids, the cluster centers are found within the data points themselves. Due to this "Medoid" strategy, *k*-medoids is less affected by noise and other outliers as compared to *k*-means clustering [18].

### 4.5 A Fast Hybrid k-Means Level Set Algorithm For Segmentation:

This algorithm [19] first draws a connection between a level set algorithm and k-Means plus nonlinear diffusion preprocessing. Then, this link is exploited to develop a new hybrid numerical technique for segmentation that draws on the speed and simplicity of k-Means procedures, and the robustness of level set algorithms. The method retains spatial coherence on initial data characteristic of curve evolution techniques, as well as the balance between a pixel/voxel's proximity to the curve and its intention to cross over the curve from the underlying energy. However, it is orders of magnitude faster than standard curve evolutions. Also, it does not suffer from the limitations of k-Means due to inaccurate local minima and allows for segmentation results ranging from k-Means clustering type partitioning leveling set partitions.

### 4.6 K-d Tree Based Efficient K-Means Algorithm:

This algorithm organizes all patterns in a k-d tree structure such that one can find all the patterns which are closest to a given prototype at the root level [20]. For the children of the root node, it is possible to prune the candidate set by using simple geometrical constraints. This approach can be applied recursively until the size of the candidate set is one for each node. Experimental results showed that this algorithm can improve the computational speed of the direct K-Means algorithm by an order to two orders of magnitude in the total number of distance calculations and the overall time of computations.

## 5. Soft Clustering Algorithms for image segmentation:

Soft clustering algorithms are based on fuzzy logic based calculations. Following are some frequently preferred soft clustering techniques used for image segmentation:

### 5.1 Fuzzy C-Means:

*Fuzzy c-means* (FCM) is a data clustering technique wherein each data point belongs to a cluster to some degree that is specified by a membership grade. This technique was originally introduced by Jim Bezdek in 1981 [21] as an improvement on earlier clustering methods [22]. This algorithm is based on minimization of the following objective function:

$$J_m = \sum_{i=1}^{N} \sum_{j=1}^{C} u_{ij}^m \|x_i - c_j\|^2$$

$$1 \leq m < \infty$$

where *m* is any real number greater than 1, $u_{ij}$ is the degree of membership of $x_i$ in the cluster *j*, $x_i$ is the *i*th of d-dimensional measured data, $c_j$ is the d-dimension center of the cluster, and ||*|| is any norm expressing the similarity between any measured data and the center. A proper value of m is very necessary for a better fuzzy clustering task since it is the factor which determines the fuzziness of the algorithm [22]. Fuzzy partitioning is carried out through an iterative optimization of the objective function shown above, with the update of membership $u_{ij}$ and the cluster centers $c_j$ by:

$$u_{ij} = \frac{1}{\sum_{k=1}^{C}\left(\frac{\|x_i - c_j\|}{\|x_i - c_k\|}\right)^{\frac{2}{m-1}}}$$

$$c_j = \frac{\sum_{i=1}^{N} u_{ij}^m \cdot x_i}{\sum_{i=1}^{N} u_{ij}^m}$$

The stopping criteria for the iterations is :

$$\max_{ij}\left\{\left|u_{ij}^{(k+1)} - u_{ij}^{(k)}\right|\right\} < \varepsilon$$

where $\varepsilon$ is a termination criterion between 0 and 1, whereas $k$ are the iteration steps. This procedure converges to a local minimum or a saddle point of $J_m$.

The formal algorithm is :

1. *Initialize U=[$u_{ij}$] matrix, $U^{(0)}$*

2. *At k-step: calculate the centers vectors $C^{(k)}$=[$c_j$] with $U^{(k)}$*

$$c_j = \frac{\sum_{i=1}^{N} u_{ij}^m \cdot x_i}{\sum_{i=1}^{N} u_{ij}^m}$$

3. *Update $U^{(k)}$, $U^{(k+1)}$*

$$u_{ij} = \frac{1}{\sum_{k=1}^{C}\left(\frac{\|x_i - c_j\|}{\|x_i - c_k\|}\right)^{\frac{2}{m-1}}}$$

4. *If $\| U^{(k+1)} - U^{(k)} \| < \varepsilon$ then STOP; otherwise return to step 2.*

FCM's performance is overall good, but the only disadvantage is that it carries a very high complexity with its calculations involved.

### 5.2 Penalty Based FCM algorithm (PFCM):

PFCM overcomes the noise sensitiveness of conventional FCM clustering algorithm [24]. This algorithm is developed by modifying the objective function of the standard FCM algorithm with a penalty term that takes into account the influence of the neighboring pixels on the centre pixels. The penalty term acts as a regularizer in this algorithm, which is inspired from the neighborhood expectation maximization algorithm and is modified in order to satisfy the criterion of the FCM algorithm. When the performance of this algorithm is discussed and compared to those of many derivatives of FCM algorithm, the experimental results on segmentation of synthetic and real images demonstrate that the algorithm is effective and robust.

### 5.3 Improved Fuzzy C Means Algorithm (IFCM):

This algorithm is based on the concept of data compression where the dimensionality of the input is highly reduced [25]. The data compression includes two steps: quantization and aggregation. The quantization of the feature space is performed by masking the lower 'm' bits of the feature value. The quantized output will result in the common intensity values for more than one feature vector. In the process of aggregation, feature vectors which share common intensity values are grouped together. A representative feature vector is chosen from each group and they are given as input for the conventional FCM algorithm. Once the clustering is complete, the representative feature vector membership values are distributed identically to all members of the quantization level. Since the modified FCM algorithm uses a reduced dataset, the convergence rate is highly improved when compared with the conventional FCM.

### 5.4 Generalized Spatial Fuzzy C Means Algorithm (GSFCM):

Medical images are inherently low contrast, vague boundaries, and high correlative [26]. The traditional fuzzy c-means (FCM) clustering algorithm considers only the pixel attributes. This leads to accuracy degradation with image segmentation. Generalized Spatial Fuzzy C-Means (GSFCM) algorithm solves this problem by utilizing both given pixel attributes and the spatial local information which is weighted correspondingly to neighbor elements based on their distance attributes. This improves the segmentation performance dramatically. Experimental results with several magnetic resonance (MR) images show that the proposed GSFCM algorithm outperforms the traditional FCM algorithms in the various cluster validity functions [26].

### 5.5 Fuzzy C-means clustering algorithm based on kernel method (FKCM):

Fuzzy kernel C-means clustering algorithm (FKCM) is based on conventional fuzzy C-means clustering algorithm (FCM). But this algorithm integrates FCM with Mercer kernel function and deals with some issues in fuzzy clustering. The properties of the new algorithms are illustrated the FKCM algorithm is not only suitable for clusters with the spherical shape, but also other non-spherical shapes such as annular ring shape effectively[27].

### 5.6 Kernel-based Fuzzy and Possibilistic C-Means Clustering:

These two algorithms substitute the original Euclidean distance by a kernel-induced distance metric for the algorithms FCM and PCM, and the corresponding algorithms are called kernel fuzzy c-means (KFCM) and kernel possibilistic c-means (KPCM) algorithms. The test results show that these two algorithms perform better than the FCM and PCM algorithms [28].

## 6. An Experiment showing color image segmentation with K-Means in Matlab:

To observe the quality of image segmentation task with clustering approach, we have performed an experiment in Matlab. For the clustering task, we have chosen K-Means algorithm. We preferred color image segmentation task. First the original image is converted from RGB color space to L*a*b space [29]. Then, we classified the colors 'a*b*' space using K-Means clustering algorithm. For K-Means clustering, we have used the kmeans function in Matlab. From the results obtained from K-Means clustering, every pixel is labeled in the image. At last, three different images are created with respect to three different clusters.

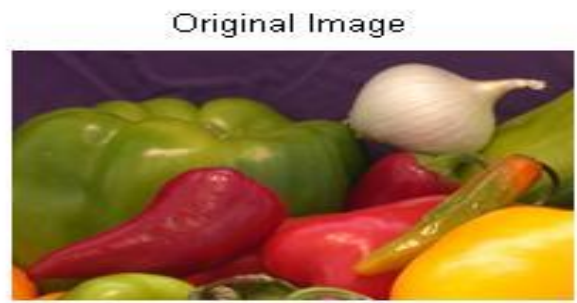

**Figure 1**: Original Image

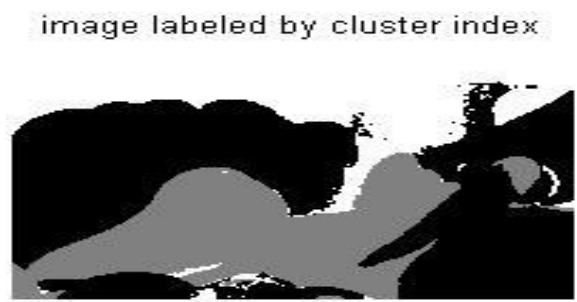

**Figure 2**: Image Obtained After K-Means Clustering

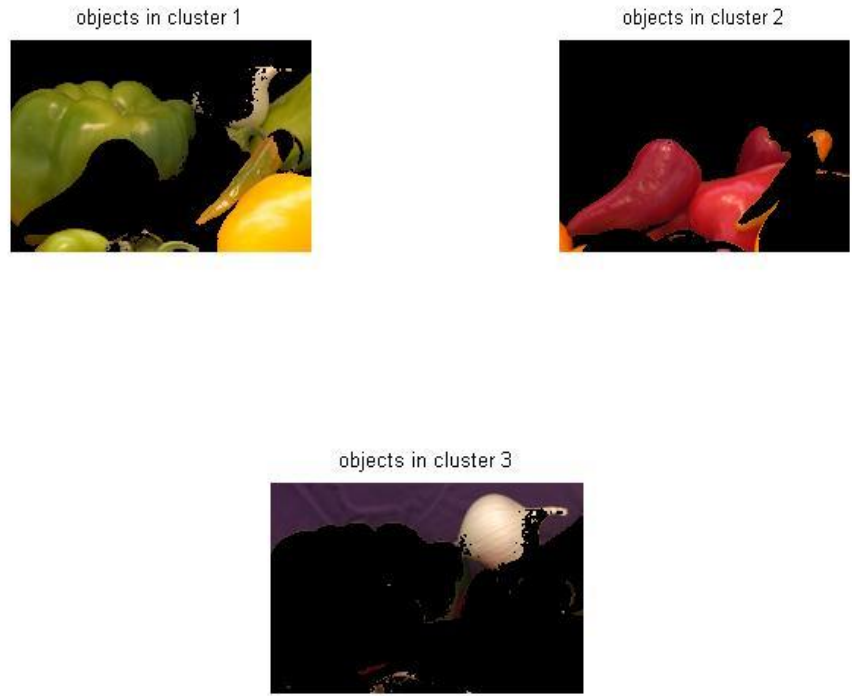

**Figure 3**: Three Different Images With Respect To Three Different Clusters

From the experiment, it is clearly visible from the segmented images, that clustering is well suited for image segmentation task. But it is really a challenging task to choose a proper clustering either hard or soft for a particular image segmentation task as it will depend on the properties of that particular image concerned.

## 7. Conclusion:

In this paper, an in-depth study is done to observe the utilization of clustering algorithms in image segmentation task. For this study, first we divide the available clustering techniques into two different categories depending on their properties and behavior as hard clustering and soft clustering techniques. Then, we picked up some of the important ones from both of these two different types and an analytical study is done on them to find out their pros and cons. Lastly, we have performed an experiment to observe the quality of output of image segmentation using clustering technique. In our future research, our main objective will be to build a novel clustering algorithm to deal with segmentation task for noisy images and also to reduce the computational cost involved in the same.

**References:**

[1] Chris Solomon, Toby Breckon ,Fundamentals of Digital Image Processing, ISBN 978 0 470 84472 4

[2] Rajesh Dass, Priyanka, Swapna Devi, "Image Segmentation Techniques", IJECT Vol.3, Issue 1, Jan-March 2012.

[3] B. Georgescu, I. Shimshoni, P. Meer, "Mean Shift Based Clustering in High Dimensions: A Texture

Classification Example", Intl Conf on Computer Vision, 2003

**A Brief Author Biography:**

**1st Author:**

*Mr. Dibya Jyoti Bora*: Ph.D. in Computer Science pursuing, M.Sc. in Information Technology (University 1st Rank holder Barkatullah University, Bhopal), GATE CS/IT Qualified, GSET qualified in Computer Science. Currently teaching PG students in the Department of Computer Science & Applications, Barkatullah University, Bhopal. Research interests are Cluster Analysis and its applications in Image Processing and Machine Learning.

**2nd Author:**

*Dr. Anil Kumar Gupta*: Ph.D. in Computer Science (Barkatullah University, Bhopal), HOD of the Department of Computer Science & Applications, Barkatullah University, Bhopal. Research interest: Data Mining, Artificial Intelligence and Machine Learning.